\documentclass{article}

\usepackage{PRIMEarxiv}

\usepackage[utf8]{inputenc} % allow utf-8 input
\usepackage[T1]{fontenc}    % use 8-bit T1 fonts
\usepackage{hyperref}       % hyperlinks
\usepackage{url}            % simple URL typesetting
\usepackage{booktabs}       % professional-quality tables
\usepackage{amsfonts}       % blackboard math symbols
\usepackage{nicefrac}       % compact symbols for 1/2, etc.
\usepackage{amsmath}
\usepackage{multirow}
\usepackage{microtype}      % microtypography
\usepackage{lipsum}
\usepackage{fancyhdr}       % header
\usepackage{graphicx}       % graphics
\usepackage{pifont}% 
\newcommand{\cmark}{\ding{51}}%
\graphicspath{{media/}}     % organize your images and other figures under media/ folder

\title{Multi-task Learning with Active Learning for
Arabic Offensive Speech Detection}
\author{
  Aisha Alansari \\
  Department of Information and Computer Science \\
  King Fahd University of Petroleum and Minerals \\
  Dhahran, Saudi Arabia\\
  \texttt{\{aisha.ansari\}@kfupm.edu.sa} 
  \AND
  Hamzah Luqman \\
  SDAIA-KFUPM Joint Research Center for Artificial Intelligence \\
  King Fahd University of Petroleum and Minerals \\
  Dhahran, Saudi Arabia\\ 
  \texttt{\{hluqman\}@kfupm.edu.sa}
}

\begin{document}
\maketitle

\begin{abstract}
The rapid growth of social media has amplified the spread of offensive, violent, and vulgar speech, which poses serious societal and cybersecurity concerns. 
Detecting such content in Arabic text is particularly complex due to limited labeled data, dialectal variations, and the language's inherent complexity.
This paper proposes a novel framework that integrates multi-task learning (MTL) with active learning to enhance offensive speech detection in Arabic social media text. By jointly training on two auxiliary tasks, violent and vulgar speech, the model leverages shared representations to improve the detection accuracy of the offensive speech. Our approach dynamically adjusts task weights during training to balance the contribution of each task and optimize performance. To address the scarcity of labeled data, we employ an active learning strategy through several uncertainty sampling techniques to iteratively select the most informative samples for model training. We also introduce weighted emoji handling to better capture semantic cues. Experimental results on the OSACT2022 dataset show that the proposed framework achieves a state-of-the-art macro F1-score of 85.42\%, outperforming existing methods while using significantly fewer fine-tuning samples. The findings of this study highlight the potential of integrating MTL with active learning for efficient and accurate offensive language detection in resource-constrained settings.
\end{abstract}

% keywords can be removed
\keywords{Offensive speech detection \and Arabic offensive speech detection \and Multi-task learning \and Uncertainty sampling \and ArabBERT}

\section{Introduction}

The rapid growth of social media platforms such as Twitter (recently changed to X), Facebook, and Instagram has revolutionized global communication. These platforms enabled users to share their ideas, opinions and information. However, this digital connectivity has also facilitated the widespread of offensive, aggressive, and harmful content. The absence of strict regulations and the anonymity provided by social media platforms often encourage users to exhibit abusive behavior, such as using offensive, violent, and vulgar language. This toxic online environment not only disrupts public discourse but also poses severe psychological and societal risks, such as emotional distress and violence, that may result in suicide cases \cite{van2018cyber}.

Abusive language encompasses a wide range of expressions that may be offensive, aggressive, hateful, violent, or vulgar. Davidson et al. \cite{davidson2017automated} distinguished between hate speech and offensive speech. Offensive language includes any form of communication that contains either implicit or explicit insults or attacks directed at others, or any language deemed inappropriate or improper. In contrast, hate speech refers to derogatory language that specifically targets individuals (such as politicians or celebrities) or particular groups (such as those defined by gender, religion, or nationality) \cite{fortuna2018survey}.

The detection of offensive speech is a critical component in mitigating the growing threat of online toxicity. While numerous studies have explored offensive language detection in English and other widely spoken languages, the Arabic language remains relatively underexplored. This gap is significant considering the increasing use of the Arabic language on social media. According to a social media analysis \cite{salem2017social}, the use of the Arabic language in social media reached an average rate of 55\% in 2017. Detecting offensive speech in Arabic poses unique challenges, such as the lack of annotated datasets, the presence of multiple dialects and writing styles, and the frequent use of context-dependent expressions and emojis \cite{al2024wasm}. These challenges increase the linguistic complexity of the task, which makes traditional detection methods less effective.

Traditional offensive speech detection approaches depend on single-task learning (STL), where models are trained independently for each task.  Although the effectiveness of STL in well-resourced settings, it fails to leverage the shared semantic and syntactic knowledge that could be gained from related tasks, such as detecting offensive speech, vulgar language, and violent threats that often co-occur in online interactions \cite{zhang2018overview}. Multi-task learning (MTL) addresses this limitation by training a single model on multiple related tasks simultaneously, allowing it to learn generalized representations that improve performance across all tasks. It assumes that all learning processes, or at least a portion of them, are related to each other \cite{standley2020tasks}.  MTL offers several advantages, including better data efficiency, reduced overfitting, and faster convergence, as auxiliary tasks provide additional learning signals that enhance the model's understanding of the primary task \cite{li2021empirical, MTL}.

Despite the advantages of MTL, its effectiveness depends on the availability of sufficiently informative data for each task \cite{yamani2024active}. To address this, active learning has emerged as a promising strategy to reduce annotation costs by selecting only the most informative samples for training \cite{dor2020active}. Among various techniques, uncertainty sampling is one of the most widely used, as it enables the model to identify samples that lack confidence and prioritize them for annotation \cite{settles2009active, pilault2020conditionally}. When combined with MTL, active learning not only enhances efficiency but also helps balance the contributions of multiple tasks and improve the overall model performance.

In this paper, we propose a novel framework that integrates MTL with uncertainty-based active learning to detect offensive speech in Arabic text. The framework is designed to jointly train on three related tasks: offensive, violent, and vulgar speech. We also propose multiple task weighting strategies to control the influence of each task during training. To address the scarcity of labeled data, we introduce an active learning strategy that uses entropy-based uncertainty measures to select the most informative training samples. A weighted emoji handling mechanism is also introduced to better capture the semantic implications of emojis, which are commonly used in Arabic online conversations.

The contributions of this paper can be summarized as follows:
\begin{itemize}
    \item Proposing an MTL framework that enhances the performance of offensive speech detection by leveraging auxiliary tasks.
    \item Using multiple task weighting techniques to optimize the training process and improve task balancing of MTL approach.
    \item Introducing an active learning strategy using entropy-based uncertainty sampling to reduce the amount of labeled data required to training the MTL model.
    \item Incorporating a weighted emoji handling mechanism that captures additional semantic cues from social media text.
    \item Achieving a new state-of-the-art performance on OSACT2022 dataset with significantly fewer training samples compared with previous approaches.

\end{itemize}

\section{Literature review}
\vspace{2mm}\noindent\textbf{STL for offensive speech detection.}
Most of the approaches that have been proposed in the literature for Arabic offensive speech are STL approaches, where only datasets related to this task are used to train the detection model. Shannaq et al. \cite{shannaq2022offensive} propose a two-stage optimization approach using XGBoost, support vector machines (SVM), and genetic algorithms to classify offensive tweets in Arabic social media. The Arabic Cyberbullying Corpus (ArCybC) was used to evaluate the proposed approach and an F1-score of 87.8\% was reported. 
Khan et al. \cite{khan2023offensive} introduced a bidirectional long-term short-term memory (BiLSTM)-based model trained on a newly constructed Roman Pashto dataset. Features extracted from BoW and TF-IDF were combined, providing a strong STL baseline in an extremely under-resourced language with an accuracy of 97.2\%. Saeed et al. \cite{saeed2023detection} provided a thorough annotation schema for Urdu, distinguishing between various offense severity levels, such as symbolization, insult, and attribution. Their work demonstrated the significance of multi-label modeling for low-resource, rich-morph languages. Niraula et al. \cite{niraula2021offensive} developed the first  Nepali offensive language dataset and evaluated STL models, including SVM and logistic regression (LR). Their findings highlight domain-specific preprocessing challenges, including mixed-script text and code-switching. Boulouard et al. \cite{boulouard2022detecting} presented a transfer learning solution using BERT-based models to classify comments as abusive or neutral. 

Besides, interpretable models have emerged to address the black-box nature of deep learning models in STL setups. Babaeianjelodar et al. \cite{babaeianjelodar2022interpretable} proposed an XGBoost-based classifier that outperformed LSTM and AutoGluon models on balanced and imbalanced Twitter datasets. They also incorporated SHAP values for model interpretability, which effectively distinguishe between hate and offensive content. Anand et al. \cite{anand2023deep} combined fuzzy convolutional neural networks (CNN)-based feature selection with a BiLSTM, Naïve Bayes, and SVM ensemble. This approach has been evaluated on a multilingual dataset collected from YouTube and Facebook, and an F1-score of 92.5\% is reported.

Recently, social information has been merged with deep contextual models to further improve STL capabilities. Miao et al. \cite{miao2023detecting} introduced GF-OLD, a novel model that combines BERT embeddings and user history and community structure via Graph Attention Networks (GANs). This hybridization of textual and social information attained a peak F1 score of 89.94\%, which indicates that social context can be employed to benefit STL even without turning to MTL architectures. 

In the context of the Arabic language, multiple studies utilized the OSACT2020 dataset \cite{husain2020osact4} to train and evaluate offensive language detection models. Elmadany et al. \cite{elmadany2020leveraging} utilized pre-trained models to detect offensive language, hence enhancing the efficiency of BERT. The proposed approach depends on several data augmentation techniques to balance the dataset. Moreover, Haddad et al. \cite{haddad2020arabic} used CNN and bidirectional gated recurrent units (BiGRU) augmented with attention mechanisms. Alharbi et al. \cite{alharbi2020combining} achieved higher results by proposing a character-level word embedding and combining it with two word-level embeddings (AraVec and Mazajak) to train an LSTM.

Socha and Kasper \cite{socha2020ks} compared monolingual and multilingual models for identifying offensive language using a fine-tuned transformer. The findings of their study show that monolingual models outperform multilingual models.  Safaya et al. \cite{safaya2020kuisail} showed that the integration of CNN and BERT can effectively improve the accuracy of offensive language detection. Husain et al. \cite{husain2020salamnet} compared the performance of several classic machine learning models with BiGRU using the OffensEval2020 dataset \cite{zampieri2020semeval}. 
 Hassan et al. \cite{hassan2020alt} demonstrated that the integration of SVM, deep neural network, and fine-tuned bidirectional encoder representations from BERT yields substantial results. Alharbi et al. \cite{alharbi2020bhamnlp} merged the SemEval2020 dataset with the OffenEval2020 \cite{zampieri2020semeval} dataset. An ensemble technique was developed by integrating multiple word embedding models and emotion transfer learning, resulting in significant outcomes.

Paula et al.'s \cite{de2022upv} used a variety of transformer models and ensemble techniques. The authors highlighted the significance of model diversity and ensemble techniques in achieving robust performance across various subtasks. The proposed approach was evaluated on the OSACT2022 \cite{mubarak2022overview} dataset. Elkaref et al. \cite{elkaref2022guct} underlined the significance of improving embedding representations, with notable performance gains coming from their cluster-based method of increasing the embedding space's isotropy. Besides, the ability of alternative loss functions to address class imbalance was shown by Mostafa et al. \cite{mostafa2022gof}, which is critical for enhancing model performance in real-world scenarios. More recently, Mousa et al. \cite{mousa2024detection} introduced a cascaded model combining ArabicBERT, BiLSTM, and RBF to categorize offensive speech into five categories using a private dataset. The cascaded model achieved an F1-score of 98.4\%.

\vspace{2mm}\noindent\textbf{MTL for Offensive speech detection.}
MTL approaches leverage shared information across related tasks to enhance the detection of offensive speech. MTL techniques can be broadly categorized into hard and soft parameter sharing \cite{ruder2019latent}. In the hard parameter sharing approach, all tasks utilize the same model or shared layers (typically the lower-level layers), while maintaining task-specific output layers. Conversely, in the soft parameter sharing approach, each task has its own model, and similarity between models is encouraged through regularization methods. Zhang and Yang \cite{zhang2018overview} further categorized parameter sharing into five types: feature learning, low-rank methods, task clustering, task relation learning, and decomposition techniques. Crawshaw et al. \cite{crawshaw2020multi} provided a comprehensive review of the optimization methods used in MTL. Additionally, Zhang et al. \cite{zhang2018overview} classified learning techniques into joint training (parallel learning) and multi-step training (sequential learning). In joint training, multiple tasks are learned simultaneously, whereas in multi-step training, tasks are learned sequentially, with the output or hidden representations from one task serving as input to the next. 

Transformer-based models have been integrated with MTL to improve performance across various tasks. For instance, Mishra et al. \cite{mishra2021exploring} proposed a single transformer-based MTL model for the HASOC 2019 competition \cite{mandl2019overview}, aimed at detecting hate and offensive speech in English, Hindi, and German. Similarly, Hande et al. \cite{hande2022multi} applied MTL to low-resourced Dravidian languages by jointly training models for offensive language detection and sentiment analysis on code-mixed YouTube comments in Tamil, Malayalam, and Kannada.

Beyond multilingual and code-mixed challenges, other studies have focused on enriching offensive language detection through auxiliary linguistic features. Plaza-del-Arco et al. \cite{plaza2022integrating} proposed an MTL model that incorporates both implicit and explicit phenomena, such as sarcasm, insults, emotional tone, and constructiveness, to enhance detection in Spanish news comments. Through the selection of suitable features with mutual information and encoding them as auxiliary tasks, their approach outperformed a strong monolingual BERT baseline, demonstrating that integrating semantic clues significantly improves detecting subtle types of offensiveness.
In another study, Zampieri et al. \cite{zampieri2023offensive} introduced a transformer model that performs post-level and token-level offense detection in conjunction with auxiliary tasks such as humor, gender bias, and engaging content detection.

For the Arabic language, Djandji et al. \cite{djandji2020multi} demonstrated the effectiveness of an MTL approach that integrates shared and task-specific layers using the OSACT2020 dataset. The authors combined offensive and hate speech detection tasks, leveraging shared characteristics between them to improve overall performance. Furthermore, Bennessir et al. \cite{bennessir2022icompass} employed the OSACT2022 dataset along with samples from L-HSAB, highlighting the benefits of MTL in enhancing classification outcomes. Shapiro et al. \cite{shapiro2022alexu} conducted a study analyzing various training strategies, such as ensemble methods, contrastive learning, and MTL, to fine-tune transformer models using multiple datasets, yielding promising results.
A summary of the Arabic surveyed techniques is shown in Table \ref{tab:LR}. Learning refers to either STL or MTL.

\begin{table}[]
\caption{Summary of surveyed offensive Arabic speech detection techniques.}
\label{tab:LR}
\resizebox{\linewidth}{!}{%
\begin{tabular}{lcclll}
\toprule
\multicolumn{1}{c}{\multirow{2}{*}{Reference}} & \multicolumn{2}{l}{Learning Method}                      & \multicolumn{1}{c}{\multirow{2}{*}{Method}} & \multicolumn{1}{c}{\multirow{2}{*}{Dataset}} & \multicolumn{1}{c}{\multirow{2}{*}{F1-score}} \\ \cline{2-3}
\multicolumn{1}{c}{}                           & \multicolumn{1}{l}{STL}    & MTL                         &                        &                         &                        \\ \midrule
Alharbi et al. \cite{alharbi2020bhamnlp}         & \cmark &                             & Ensemble Learning                            & OffensEval2020                                & 0.870                                          \\  
Husain et al. \cite{husain2020salamnet}          & \cmark &                             & GRU                                          & OffensEval2020                                & 0.830                                          \\  
Hassan et al. \cite{hassan2020alt}               & \cmark &                             & Ensemble Learning                            & OffensEval2020                                & 0.902                                          \\  
Socha et al. \cite{socha2020ks}                  & \cmark &                             & BERT-Base                                    & OffensEval2020                                & 0.890                                          \\  
Safaya et al. \cite{safaya2020kuisail}           & \cmark &                             & BERT+CNN                                     & OffensEval2020                                & 0.897                                          \\  
Mubarak et al. \cite{mubarak2023emojis}          & \cmark &                             & QARiB                                        & OSACT4                                        & 0.823                                          \\  
Haddad et al. \cite{haddad2020arabic}            & \cmark &                             & GRU                                          & OSACT4 + YouTube comments                     & 0.859                                          \\  
Alharbi et al. \cite{alharbi2020combining}       & \cmark &                             & LSTM                                         & OSACT4                                        & 0.868                                          \\  
Elmadany et al. \cite{elmadany2020leveraging}    & \cmark &                             & BERT-Base                                    & OSACT4                                        & 0.829                                          \\  
Mousa et al. \cite{mousa2024detection}           & \cmark &                             & Cascade model                                & Private dataset                               & 0.984                                          \\  
Paula et al. \cite{de2022upv}                    & \cmark &                             & AraBERT                                      & OSACT2022                                     & 0.827                                          \\  
Elkaref et al. \cite{elkaref2022guct}            & \cmark &                             & MARBERT                                      & OSACT2022                                     & 0.791                                          \\  
Mostafa et al. \cite{mostafa2022gof}             & \cmark &                             & Ensemble Learning                            & OSACT2022                                     & 0.852                                          \\  
Shapiro et al. \cite{shapiro2022alexu}           & \cmark &                             & MARBERT                                      & OSACT2022 + Others                            & 0.841                                          \\  
Ben Nessir et al. \cite{bennessir2022icompass}    &        & \cmark & MARBERT with QRNN                            & OSACT2022 + L-HSAB                            & 0.839                                          \\ 
Djandji et al. \cite{djandji2020multi}           &        & \cmark & AraBERT                                      & OSACT4                                        & 0.900                                          \\ \bottomrule
\end{tabular}
}
\end{table}

\section{Proposed framework}\label{sec3}

\begin{figure}[hbt!]
    \centering
    \includegraphics[width=\linewidth]{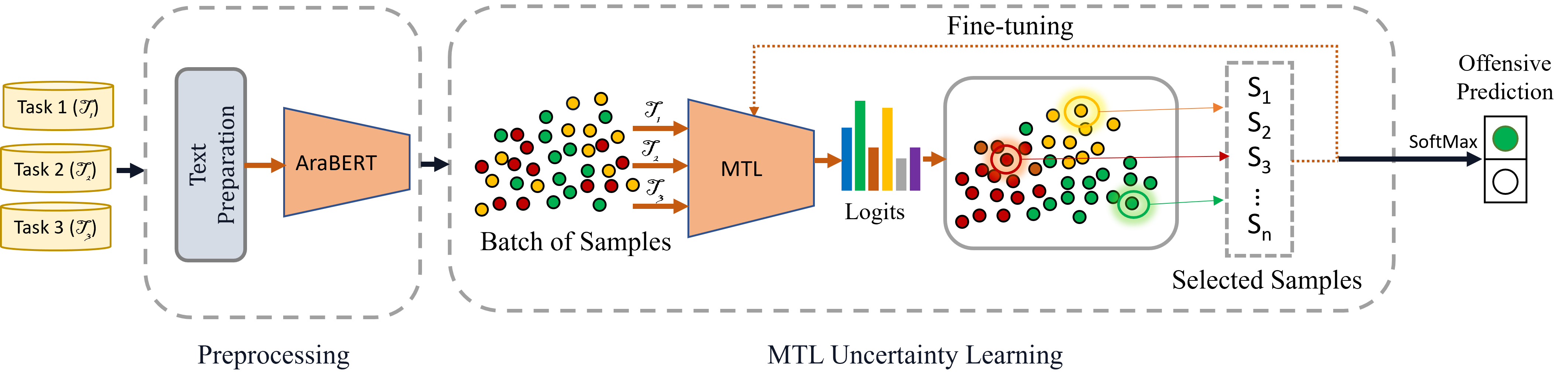} % Adjust width as needed
    \caption{The Proposed Framework.}
    \label{fig:proposed_framwork}
\end{figure}

In this work, we propose an MTL framework integrated with an active learning strategy to detect offensive speech in Arabic text. The proposed framework, shown in Figure \ref{fig:proposed_framwork}, consists of four main components: input text preprocessing, samples selection, MTL, and classification. 

\subsection{Data preprocessing}\label{subsec3.1}
The OSACT2022 dataset \cite{mubarak2023emojis} is used in this work to train and evaluate the proposed models. The dataset contains text collected from social media, particularly the X platform (previous Twitter). The dataset was collected to identify hate and offensive Arabic speech on social media. Each tweet consists of four labels: Offensive, hate, vulgar, and violent speech. In this study, we only considered offensive, vulgar, and violent speech, because in this dataset, hate speech was not independently labeled. It is only annotated after tweets are classified as offensive, which violates the assumption of task independence commonly required for MTL frameworks. 

A series of steps have been followed to prepare the dataset for training our models. The preprocessing pipeline starts with removing URLs and user mentions, and replacing tabs and long spaces with a single space. We also cleaned the text from non-textual data, such as digits and numbers. Moreover, words concatenated by special characters, such as hashtags and underscores, are separated by a space. Repeating characters, which are frequently seen in social media text to add emotion, emphasis, and personality to their messages, are condensed into a single character.  
Another level of preprocessing was performed by removing Arabic diacritics and elongation (tatweel) characters. Diacritical marks are called Tashkeel in Arabic and are used as phonetic guides, including the short vowels and the grammatical markers. They are used mainly in classic Arabic and are not heavily used in dialect and modern Arabic scripts used in the social media. 

The preprocessed text of each task is fed into an AraBERT-TwitterV2, a transformer-based language model \cite{antoun2020arabert}.  
This model is a variant of the AraBERT model that was developed to comprehend and generate Arabic text. 
AraBERT consists of 12 attention layers, 12 attention heads, 768 hidden dimensions, and a 512 maximum sequence length. 
This model was pre-trained with the same methodology as the BERT model, using masked language modeling and next-sentence prediction. In order to make sure that AraBERT accurately captures the subtleties and complexities of the Arabic language, it was pre-trained on an extensive corpus of Arabic text, which includes news articles, Wikipedia, and other publicly accessible datasets. It was trained using 2.5 billion Arabic tokens from a variety of domains, mostly using modern standard Arabic (MSA). Because of its extensive training, the model can understand various Arabic dialects and styles, which improves its performance in text classification tasks. However, in this work, we are targeting an Arabic text on social media that is usually written in the Arabic dialect language, which is different from the MSA used in AraBERT. To address this issue, we used the AraBERT-Twitter version of AraBERT, which was trained on data collected from social media, particularly Twitter. According to Alturayeif et al. \cite{alturayeif2021fine}, BERT-based models trained on social media text outperformed other variants of BERT trained on classic or MSA Arabic text.    

\subsection{Uncertainty sampling}\label{subsec3.2}
In this work, we hypothesize that the quality of training samples plays a vital role in the accuracy of the offensive speech detection model using an MTL.  Active learning has proven effective in prioritizing sample selection for labeling during training. \cite{siddhant2018deep,saeed2024active,ambati2012active}. Therefore, we used active learning to select the most informative samples from the training data. Prioritizing sample selection in the literature depends on a single task, whereas our approach is multi-task active learning, where samples are selected based on the importance of all tasks associated with those samples.

We propose utilizing an uncertainty sampling technique \cite{nguyen2022measure} in conjunction with MTL to identify the most informative samples for model training. The acquisition function, which acts as the scoring mechanism during sampling, prioritizes samples in which the classifier exhibits the highest uncertainty. These uncertain instances are likely to be the most challenging samples for classification, often situated near class boundaries. By focusing on such samples, the model gains the most insight into these critical decision regions. Leveraging this principle, we employ uncertainty sampling to select the least certain samples to be used to fine-tune AraBERT-TwitterV2 model for offensive speech detection.

Entropy sampling is used in this work to measure the uncertainty of the predictions of the AraBERT-TwitterV2 model for a given sample. Entropy quantifies the amount of uncertainty in the predicted probability distribution over the possible classes. The entropy of each class is defined as:

\begin{equation}
\mathcal{H}(x) = -\sum_{y \in Y} P(y \mid x) \log P(y \mid x)
\end{equation}

where \(p(y\mid x\)) is the model's predicted probability of a sample x for a class \(y\). The probability of each task is computed by converting the logits resulting from the model prediction into probabilities using the Sigmoid function. We computed the entropy of each task (offensive, violent, and vulgar) separately. However, the selection of each sample depends on the entropy of the three tasks of that sample. Therefore, we propose equal, weighted, and dynamic methods to combine the tasks' uncertainties of each sample.   

\begin{itemize}
    \item \textbf{Equal Entropy Sampling.} This method calculates a sample's uncertainty using the mean uncertainty of all its associated tasks' uncertainties. The mean of all tasks' uncertainties is computed as follows: 

 \begin{equation}
    \mathcal{H}(x) = \frac{1}{3} (\mathcal{H}_{\text{off}} + \mathcal{H}_{\text{vio}} + \mathcal{H}_{\text{vul}}) 
 \end{equation}

where \(\mathcal{H}_{\text{off}}\), \(\mathcal{H}_{\text{vio}}\), and \(\mathcal{H}_{\text{vul}}\) are the uncertainties of the offensive, violent, and vulgar tasks, respectively. 

 \item \textbf{Weighted Entropy Sampling. }
This method assigns different weights for each task's uncertainty. The weighted entropy is computed as follows:
\begin{equation}
    \mathcal{H}(x) = 
    \frac{
        \mathcal{W}_{\text{off}} \mathcal{H}_{\text{off}} + \mathcal{W}_{\text{vio}} \mathcal{H}_{\text{vio}} + \mathcal{W}_{\text{vul}} \mathcal{H}_{\text{vul}}
    }
    {\mathcal{W}_{\text{off}} + \mathcal{W}_{\text{vio}} + \mathcal{W}_{\text{vul}}}
\end{equation}

where \(\mathcal{W}_{off}\), \(\mathcal{W}_{vio}\), and \(\mathcal{W}_{\text{vul}}\) are the weights of the offensive, violent, and vulgar tasks, respectively.
A higher weight is given to the offensive speech task, since it is the main task in our work, while other tasks are auxiliary tasks to enhance the detection accuracy of offensive speech. Therefore, we empirically set \(\mathcal{W}_{off}\) to 2, while \(\mathcal{W}_{vio}\) and \(\mathcal{W}_{\text{vul}}\) are set to 1 each.

 \item \textbf{Dynamic Entropy Sampling. }
This technique assigns a relative weight to each task based on the overall performance of the MTL model. Compared to the weighted approach that defines a fixed weight for each task, the focus weights in the dynamic approach are modified based on the task performance (the macro F1-scores) at the end of each training iteration. The main aim of this technique is to give more priority to the task whose performance requires more improvement, which is offensive speech detection. As a result, the initial weights for vulgar, violent, and offensive tasks are set to 1/2, 2/3, and 2, respectively. After each epoch, the focus weights are adjusted based on the performance of the model in predicting the offensive speech using the following equation:

\begin{align}
\mathcal{W}_{\text{off}} = 
\begin{cases} 
\min\left( \mathcal{W}_{\max}, (\mathcal{T}_{\min} - F1_{\text{off}}) + 1.0 \right) & \text{if } F1_{\text{off}} < \mathcal{T}_{\min} \\
\max\left( \mathcal{W}_{\min}, 1.0 - (F1_{\text{off}} - \mathcal{T}_{\min}) \right) & \text{if } F1_{\text{off}} \geq \mathcal{T}_{\min}
\end{cases}
\end{align}

where \(\mathcal{T}_{min}\) is the minimum acceptable macro F1-score, set to 0.75. \(\mathcal{W}_{min}\) and \(\mathcal{W}_{max}\) are the lower and upper weight bounds set to 0.5 and 2.0, respectively. We empirically set the weight for the violent task (\(\mathcal{W}_{vio}\)) to one-third of the dynamic weight calculated for the offensive task, whereas we set the weight for the vulgar task (\(\mathcal{W}_{vul}\)) to half of the dynamic weight calculated for the offensive task as defined in this equation:

 \begin{equation}
    \mathcal{H}(x) = \mathcal{W}_{off}\mathcal{H}_{\text{off}} + \frac{2}{3}\mathcal{W}_{off}\mathcal{H}_{\text{vio}} + \frac{1}{2}\mathcal{W}_{off}\mathcal{H}_{\text{vul}} 
 \end{equation}

\end{itemize}

\subsection{Multi-task learning} \label{subsec3.3}

\begin{figure}[hbt!]
    \centering
    \includegraphics[width=\linewidth]{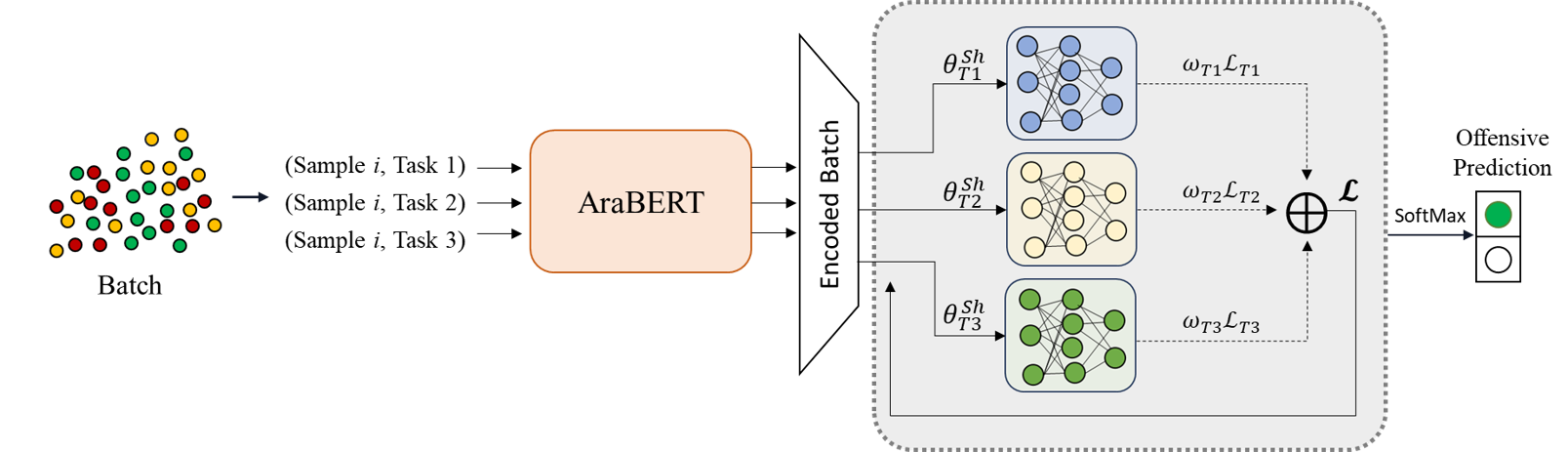} % Adjust width as needed
    \caption{The overall architecture of the MTL component. }
    \label{fig:PMTL_framwork}
\end{figure}

The main component of our proposed approach is the MTL component that handles several tasks concurrently to leverage knowledge from related tasks. MTL is useful for addressing data shortages and boosting the model's performance. Figure \ref{fig:PMTL_framwork} illustrates the overall structure of the proposed MTL. 
A batch of multi-labeled samples, each annotated for offensive, violent, and vulgar content, is fed into the AraBERT-TwitterV2 encoder to generate contextual embeddings. This model is fine-tuned using joint training and hard parameter sharing.
The resulting embeddings are fed into the MTL component, which consists of two parts: shared and specific layers. The first part, shared layers, depends on fine-tuning the AraBERT-TwitterV2 model by combining the losses of each task to learn shared information. 
The second part is the task-specific part, where a fully connected layer is dedicated to each task. The embedding of each task resulting from the encoder component is fed in parallel into the task-specific layers. The loss of each task is weighted, and the overall loss is computed. This process is iteratively repeated during fine-tuning to enhance the detection accuracy of the offensive task. The final logits after training the model are fed into a SoftMax function to select the correct class of the offensive task. Three weighting techniques have been proposed in this work to compute the overall loss of the MTL model: equal, static, and dynamic weighting. 

\begin{itemize}
    \item \textbf{Equal Weighting.} In this approach, all tasks are assigned equal weights during training. Then, a cumulative sum of the losses from each task is calculated to derive the total loss. This technique ensures that all tasks contribute equally to the optimization process, promoting uniform learning across the three tasks.

 \item \textbf{Static Weighting. } In this approach, a fixed weight is assigned to each task before training begins. The loss value of each task is multiplied by its respective weight, and the weighted losses are summed to compute the overall loss. The offensive, violent, and vulgar tasks are empirically assigned 0.7, 0.15, and 0.15 weights, respectively.  

 \item \textbf{Dynamic Weighting. } Unlike the previous methods, the dynamic weighting approach adapts tasks' weights in real-time based on the loss of the corresponding task during training.  Tasks with higher loss values are assigned more weights, which encourages the model to allocate more resources to harder tasks.

\end{itemize}

\section{Experimental work}

\subsection{Dataset}
The OSACT2022 dataset \cite{mubarak2023emojis} is used in this work to train and evaluate the proposed models.  The dataset was collected to identify hate speech and offensive Arabic language on social media, particularly Twitter. Emojis are used as anchors in the collection process of this dataset to identify tweets that are more likely to contain offensive content. The dataset consists of four classes: offensive, hate, vulgar, and violent speech. In this study, we only considered offensive, vulgar, and violent speech. Hate speech was excluded since it was not independently labeled. It is only annotated after tweets are classified as offensive. This hierarchical dependency violates the assumption of task independence commonly required for MTL frameworks. Table \ref{tab:dataset_stat} outlines the number of samples of each class in the train, development, and test sets. As shown in the table, the distribution shows that the dataset is highly imbalanced, where most samples are labeled "not violent" or "not vulgar".  

\begin{table}[h!]
    \centering
    \caption{The statistics of the OSACT2022 dataset}
    \label{tab:dataset_stat}
    \begin{tabular}{llccc}
    \toprule
Task                      & Label         & Train           & Dev           & Test           \\ \toprule
\multirow{2}{*}{Offensive} & Offensive     & 3,066           & 403            & 887            \\
                           & Not Offensive & 5,491           & 863            & 1,654          \\ \midrule
\multirow{2}{*}{Vulgar}    & Vulgar        & 132             & 16             & -              \\
                           & Not Vulgar    & 8,425           & 1,250          & -              \\ \midrule
\multirow{2}{*}{Violent}   & Violent       & 60              & 6              & -              \\
                           & Not Violent   & 8,497           & 1,260          & -              \\ \bottomrule
\multicolumn{2}{c}{\textbf{Total}}         & \textbf{8,557} & \textbf{1,266} & \textbf{2,541} \\ \bottomrule
\end{tabular}
\end{table}

\subsubsection{Experimental setup}\label{subsubsec2}

The framework was implemented using PyTorch, and the models were trained using an NVIDIA RTX A6000 GPU with 48 GB of memory and a batch size of 64.
The optimizer AdamW was used with a learning rate of 2e-5 and the BCEWithLogitsLoss was used as a loss function for all tasks. An early stopping was set with a patience of 3 to reduce the possibility of overfitting. The patience counter is incremented based on the macro F1-score of the offensive speech class. When early stopping is triggered, the best model is reloaded for testing. The macro F1-score was utilized to evaluate all the experiments. The random state was set to 42.

\subsection{Results and discussion}

We started our experiments with an STL model where only the offensive task is considered. We evaluated the performance of the STL model with and without emojis with different numbers of selected samples to fine-tune the ArabBERT model, and the obtained results are shown in Table \ref{tab:Single}. 
The results indicate that increasing the number of selected samples improves the macro F1-score, however, the improvements become marginal beyond a certain point. For instance, when excluding the emojis, the score rises from 82.72\% with 10 selected samples to 83.89\% when all samples are used.
The impact of emoji inclusion is also evident in the results. When emojis are included, the performance remains relatively stable across different sample sizes, reaching the highest F1-score of 83.89\%. However, when emojis are considered, the performance rises and an F1-score of 85.31\% is obtained, when all samples are used.
Assigning more weight to a predefined list of emojis commonly associated with offensive speech improved the model's performance compared to excluding emojis entirely, achieving an F1-score of 84.84\% when only 20 samples in each batch were utilized in each epoch. This suggests that weighting emojis helps in balancing their effect in offensive speech detection.

\begin{table}[h!]
    \centering
    \caption{The performance of the proposed approach with the STL model. The total fine-tuned samples represents the cumulative number of samples used during training, including repeated instances across epochs.}
    \label{tab:Single}
\begin{tabular}{llcccc}
\toprule
\multicolumn{3}{c}{Emoji}                                          & \multicolumn{1}{l}{\multirow{2}{*}{\# Selected Samples}} & \multicolumn{1}{l}{\multirow{2}{*}{Total Fine-tuned Samples}} & \multicolumn{1}{l}{\multirow{2}{*}{Macro F1-score}} \\ \cline{1-3}
\multicolumn{1}{c}{No}    & \multicolumn{1}{c}{Yes}     & Weighted &                                     &                                          &                                \\ \toprule 
                    \cmark       &  &          & 10                                                       & 5,360                                                         & 82.72                                              \\
                      \cmark     &   &          & 20                                                       & 5,360                                                        & 82.61                                               \\
                       \cmark    &   &          & 30                                                       & 4,020                                                        & 82.79                                               \\
                       \cmark    &   &          & 40                                                       & 16,080                                                       & 82.52                                               \\
                       \cmark    &   &          & All                                                      & 8,557                                                        & 83.89                                               \\ \midrule 
 &          \cmark                  &          & 10                                                       & 13,400                                                        & 83.69                                               \\
 &        \cmark                    &          & 20                                                       & 8,040                                                        & 83.71                                               \\
 &        \cmark                    &          & 30                                                       & 16,080                                                        & 83.20                                               \\
 &         \cmark                   &          & 40                                                       & 10,720                                                        & 83.32                                               \\
 &        \cmark                    &          & All                                                      & 17,114                                                        & 85.31                                               \\ \midrule
                           &                            & \cmark   & 10                                                       & 4,020                                                         & 84.83                                               \\
                           &                            & \cmark   & 20                                                       & 5,360                                                        & 84.84                                               \\
                           &                            & \cmark   & 30                                                       & 8,040                                                        & 84.30                                      \\
                           &                            & \cmark   & 40                                                       & 16,080                                                       & 83.82                                               \\
                           &                            & \cmark   & All                                                      & 17,114                                                       & 85.13                                               \\ \bottomrule
\end{tabular}
\end{table}

We extended our experiments to MTL with different loss weighting techniques and the obtained results are presented in Table \ref{tab:MTL}. We performed equal, static, and dynamic loss weighting across tasks with four uncertainty strategies (more details about MTL techniques and uncertainty strategies are available in Section 3.2. Based on the results obtained with the STL approach, we are reporting the performance of the MTL with weighted emojis and only 10 samples per batch selected for fine-tuning. The results with other configurations will be reported within the ablation study section.   

As shown in Table \ref{tab:MTL}, all MTL approaches improved the performance of the proposed models compared with the STL using 10 selected samples. The highest F1 scores were obtained using the dynamic weighting approach, while the lowest scores were obtained using an equal MTL weighting approach. These results align with other studies \cite{alturayeif2023enhancing} where the main task, offensive speech detection, should be given more weight compared with other supporting tasks.

For uncertainty sampling strategies, it has been noticed that assigning the offensive task more uncertainty weight results in low scores with almost all MTL approaches. In contrast to the weighted MTL approach, giving higher uncertainty weight to the offensive speech task does not improve results compared with other strategies. Additionally, using a dynamic uncertainty strategy with the dynamic weighting of MTL resulted in the lowest score compared with other strategies. The highest F1 score was obtained with an average uncertainty sampling and dynamic weighted MTL approach.  
\begin{table}[]
\centering
    \caption{The performance of the MTL techniques with different uncertainty strategies. }
    \label{tab:MTL}
\begin{tabular}{llc}
\toprule
MTL   Approach            & Uncertainty & Macro F1    \\ \toprule
\multirow{4}{*}{Equal}    & None         & 83.97 \\
                          & Equal        & 84.15 \\
                          & Weighted       & 85.07 \\
                          & Dynamic      & 83.82 \\ \midrule
\multirow{4}{*}{Static} & None         & 83.97 \\
                          & Equal        & 85.19 \\
                          & Weighted       & 83.66 \\
                          & Dynamic      & 84.57 \\ \midrule
\multirow{4}{*}{Dynamic}  & None         & 84.01 \\
                          & Equal        & \textbf{85.42} \\
                          & Weighted       & 84.60  \\
                          & Dynamic      & 84.00   \\ \bottomrule
\end{tabular}
\end{table}

While the model demonstrated strong overall performance, misclassifications reveal specific challenges related to understanding context, cultural nuances, sarcasm, emoji usage, and potential labeling inconsistencies within the dataset. 
One major issue is the misinterpretation of context, particularly in tweets where the intent to offend is inferred rather than clearly expressed. For instance, the offensive tweet \raisebox{-0.2\height}{\includegraphics[height=1em]{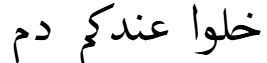}}. \raisebox{-0.2\height}{\includegraphics[height=1em]{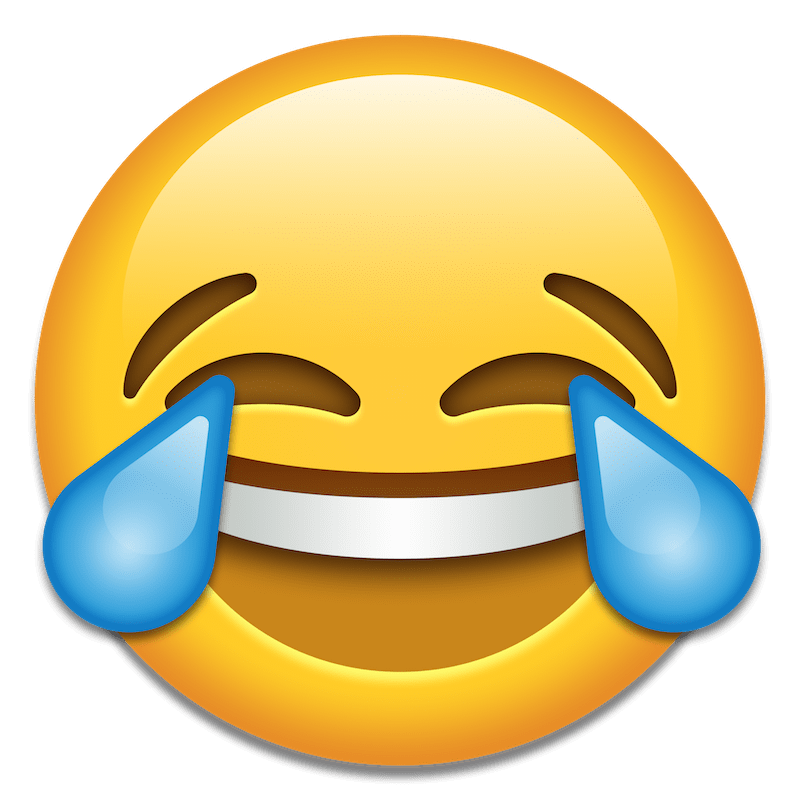}} \raisebox{-0.2\height}{\includegraphics[height=1.3em]{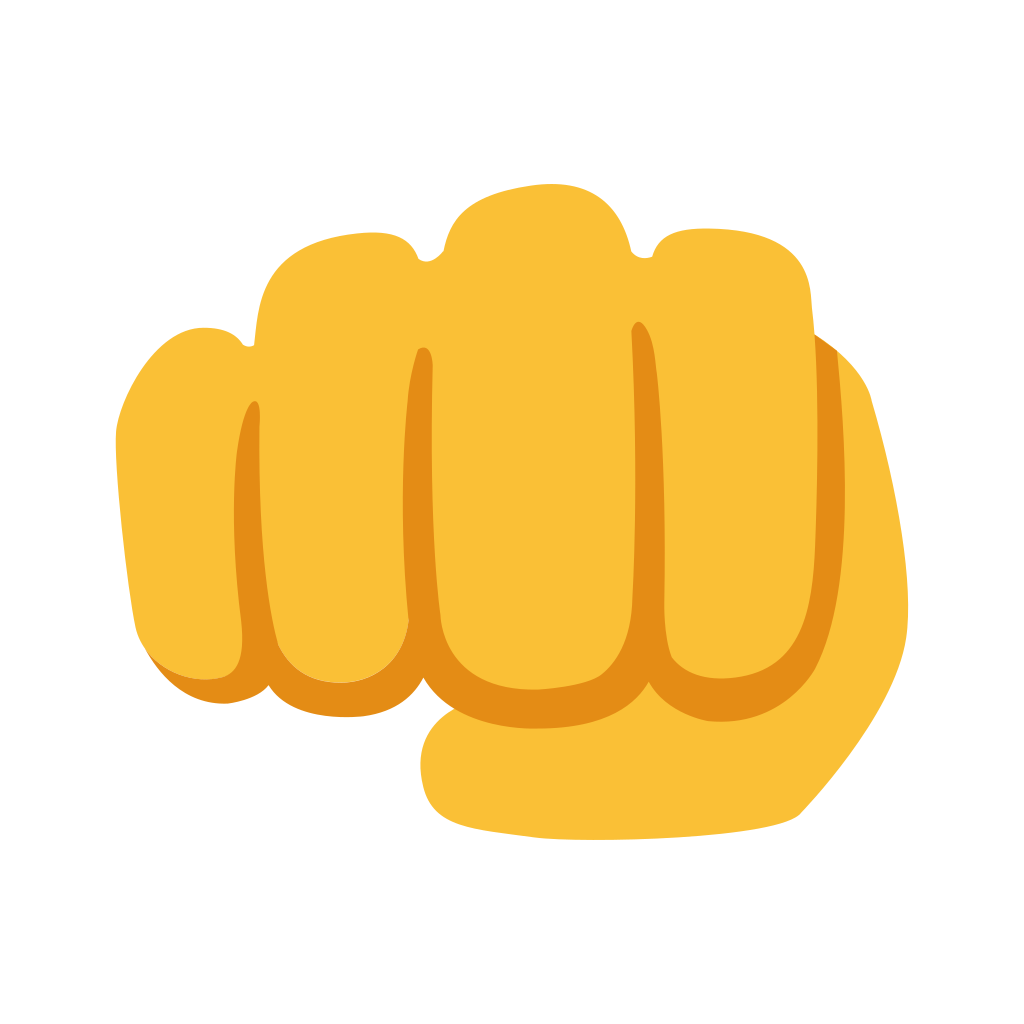}} "Have some shame"
is wrongly classified as "Not Offensive". The sarcastic tone emphasizes the offensive intent, which the model failed to detect. In another example, \raisebox{-0.2\height}{\includegraphics[height=1em]{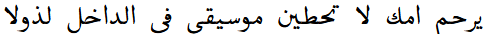}} \raisebox{-0.4\height}{\includegraphics[height=1.5em]{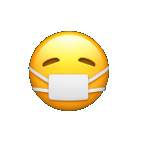}} "Please, for the love of God, don’t add the music of Fi Al-Dakhel for them", the model wrongly predicted it as "Offensive". The phrase "Fi Al-Dakhel" refers to a Turkish TV series (Fi Al-Dakhil), whose music became widely recognized and used in a variety of action contexts. The tweet, while expressing dissatisfaction with the music choice, lacks explicit offensive language. This highlights the model’s struggle to understand cultural references and contextual subtleties in Arabic tweets.

Another common error arises from the failure to recognize sarcasm or humor, where tweets use indirect or sarcastic language to convey offensive content. For instance, the tweet \raisebox{-0.2\height}{\includegraphics[height=1em]{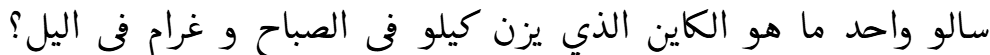}} "What is the creature that weighs a kilogram in the morning and only a gram at night?" was labeled "Offensive," but the model predicted it as "Not Offensive." The tweet’s punchline involves a culturally offensive joke, highlighting the model’s inability to detect implicit insults embedded within jokes or riddles.
Additionally, misinterpretation of context is often compounded by the incomplete nature of tweets, which may lack sufficient standalone information. Many collected tweets in the used dataset are written as replies or as part of a broader conversation. 
Therefore, to accurately annotate a tweet, the full context must be known. For example, the tweet \raisebox{-0.2\height}{\includegraphics[height=1em]{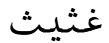}} \raisebox{-0.2\height}{\includegraphics[height=1.0em]{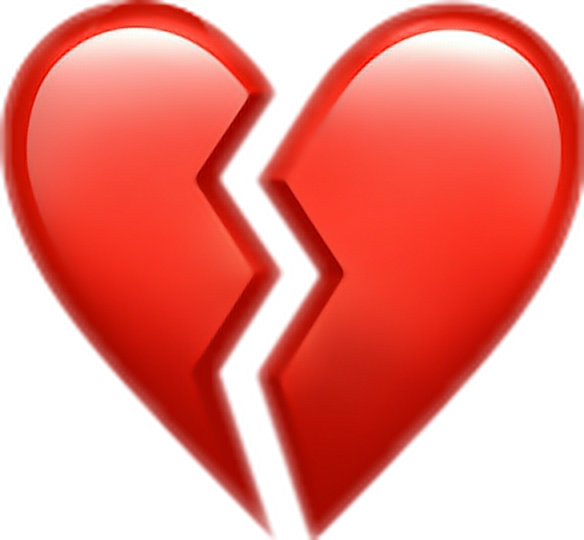}}\raisebox{-0.2\height}{\includegraphics[height=1.0em]{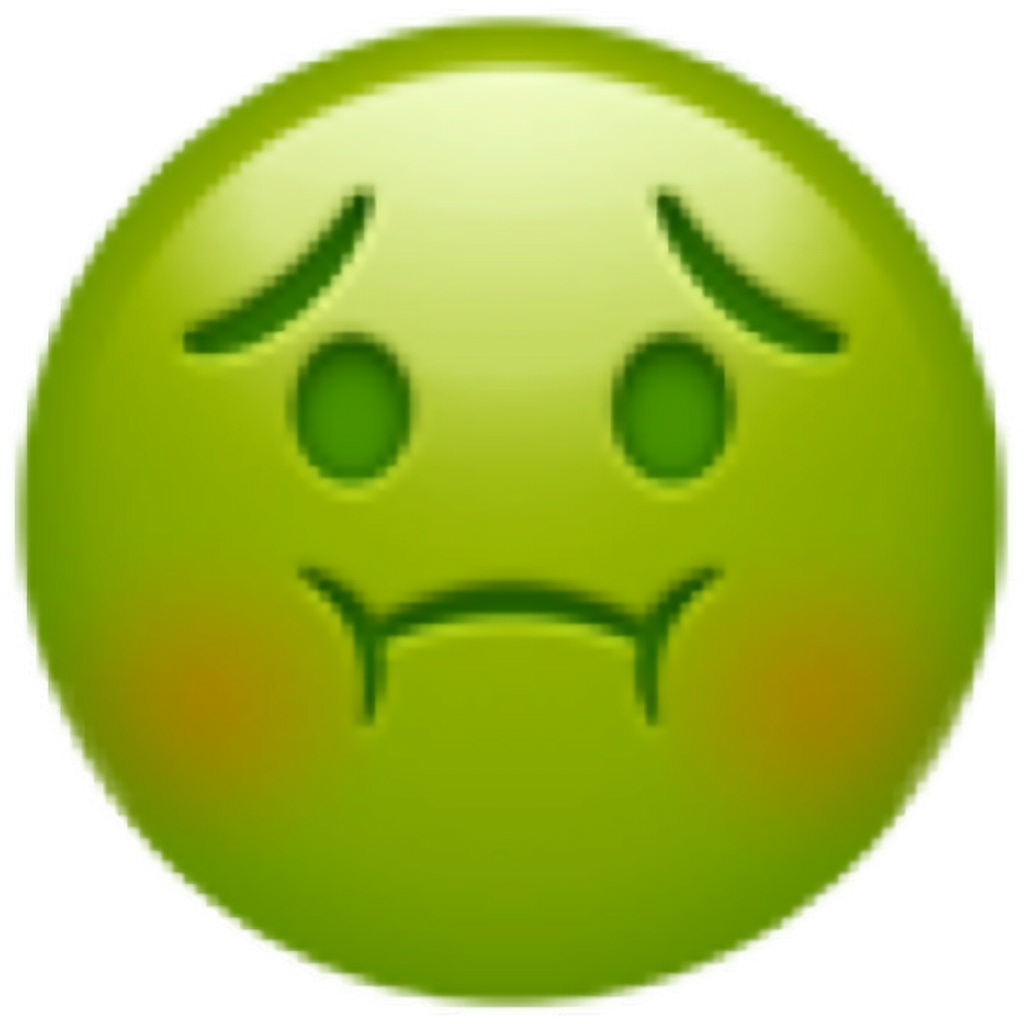}} "Annoying" was classified by the model as "Offensive," while the actual label is "Not Offensive." Similarly, in the tweet\raisebox{-0.2\height}{\includegraphics[height=1em]{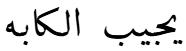}}\raisebox{-0.2\height}{\includegraphics[height=1.1em]{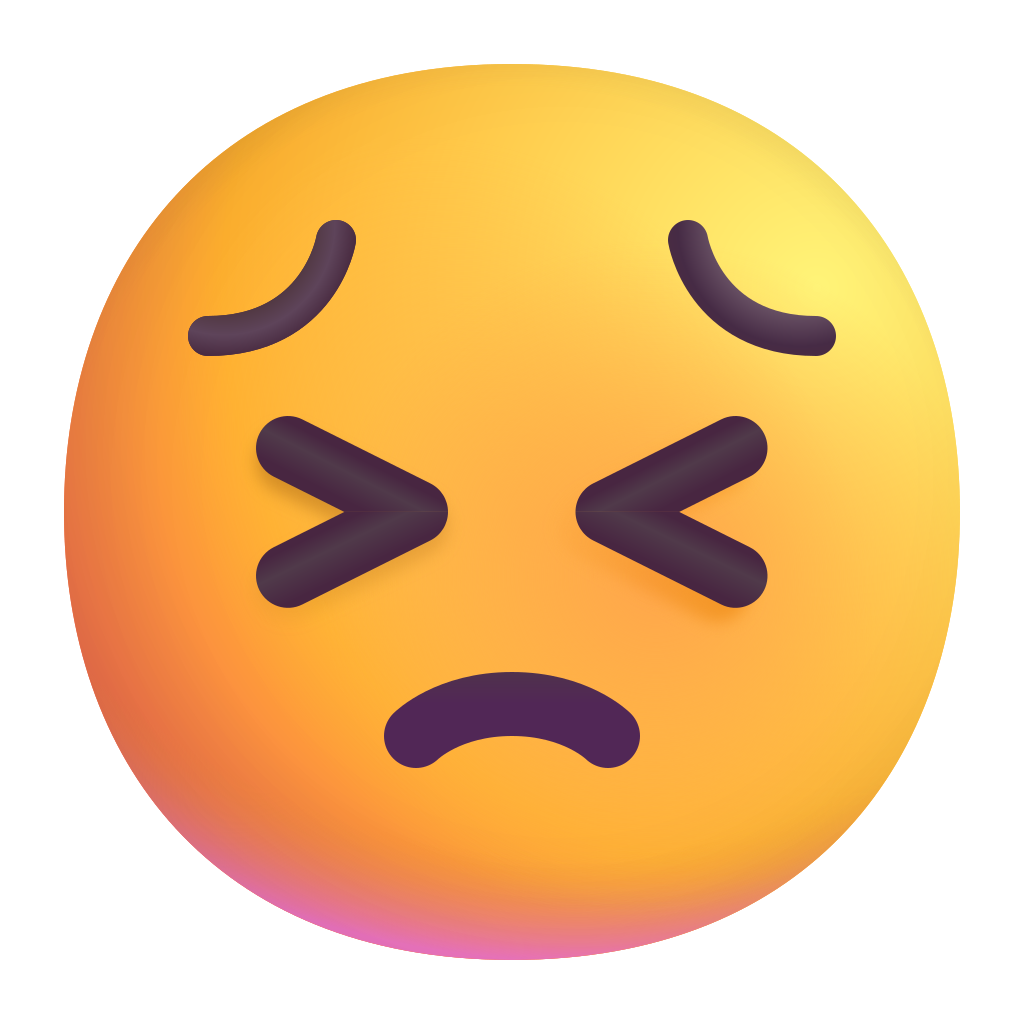}}\raisebox{-0.2\height}{\includegraphics[height=1.0em]{Images/heart.png}} "Such a mood killer", the model predicted it as "Not Offensive". These tweets do not explicitly provide offensive content and could be interpreted as offensive depending on the context they were replying to. The subjective nature of these tweets and their reliance on the preceding or related conversation create significant challenges for the offensive speech detection models, as the full intent and tone can only be understood within the larger conversation thread. This reliance on context illustrates potential errors in labeling and highlights the complexity of annotating offensive language in isolated tweets. 

Another challenge arises from using some emojis that reveal offensive or aggressive sentiments in tweets labeled as "Not Offensive." The tweet \raisebox{-0.2\height}{\includegraphics[height=1em]{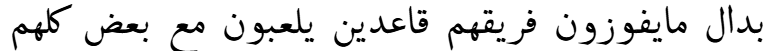}}  \raisebox{-0.2\height}{\includegraphics[height=1.0em]{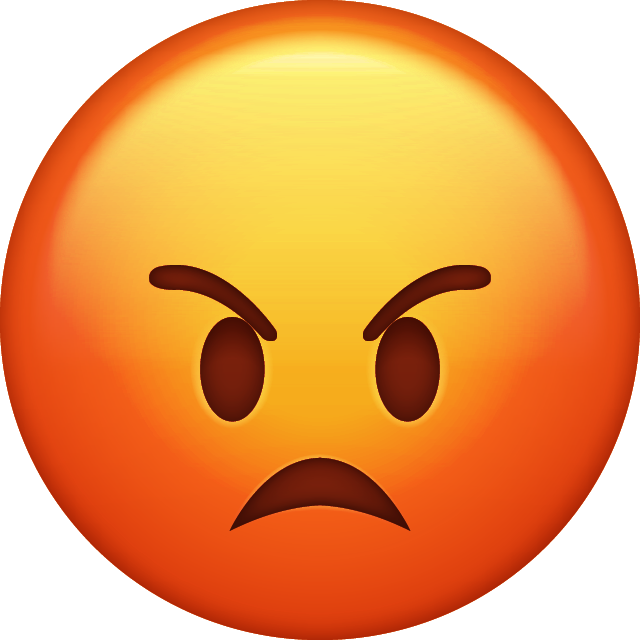}} \raisebox{-0.2\height}{\includegraphics[height=1.3em]{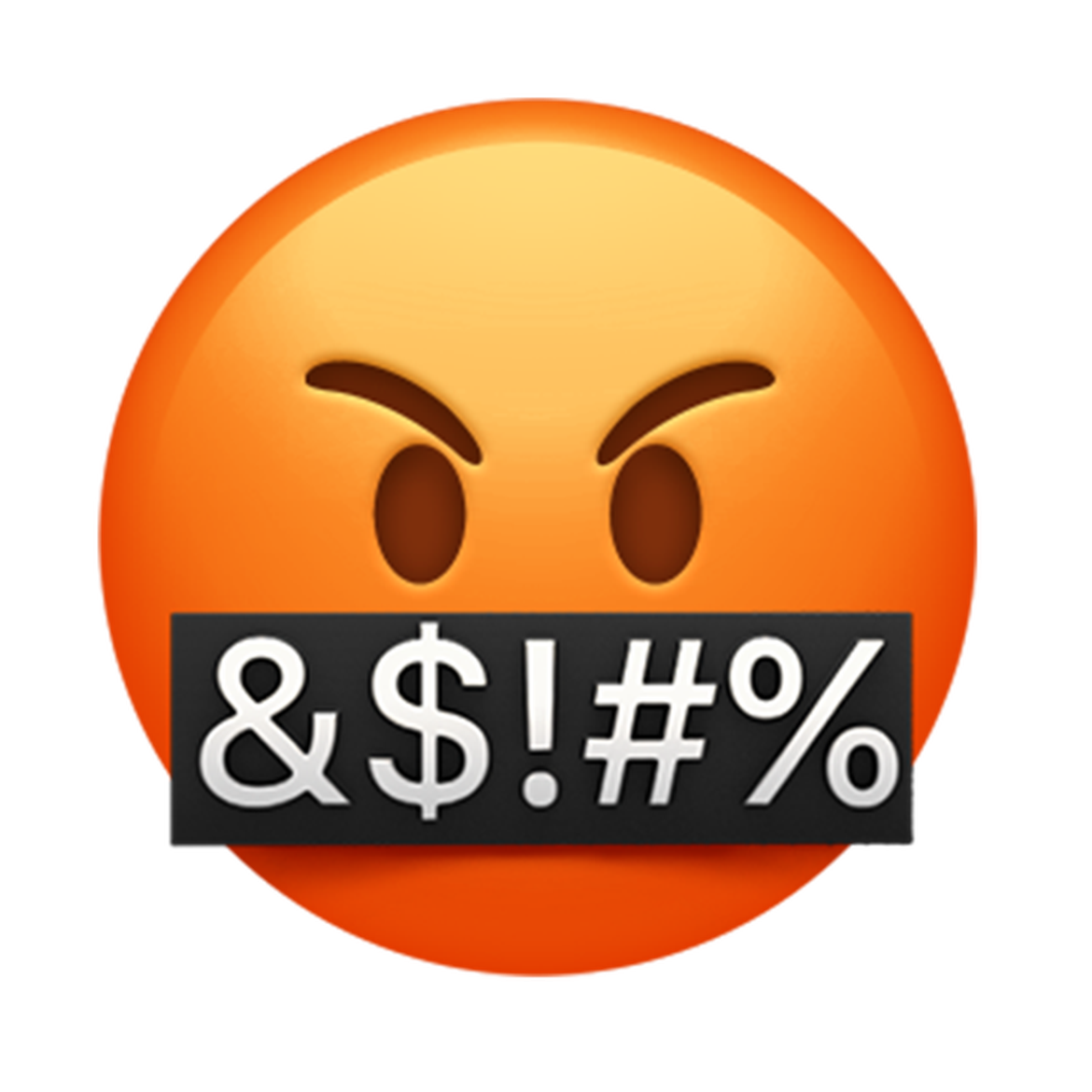}} "Rather than winning their team, they are just messing around with each other" 
%Similarly, in the tweet \RL{جوك}\raisebox{-0.15\height}{\includegraphics[height=1.0em]{sections/Poop-Emoji.png}} "Your vibe". 
 was labeled as "Not Offensive", while the model classified it as an offensive sample due to the emojis used in this tweet.  This overlap in emoji usage between offensive and non-offensive tweets highlights a nuanced challenge for models attempting to distinguish intent and sentiment based on emojis alone, which is a trade-off in using weighted emojis. 

\subsection{Comparison with other works}

We compare the performance of the proposed framework with other works in Table \ref{tab:comparison}. As shown in the table, our approach outperformed all other methods in the literature. A macro F1-score of 85.42\% was achieved using our approach with significantly fewer samples. This performance outperformed Mostafa et al. \cite{mostafa2022gof} approach that utilized the whole dataset for model training. In addition, their method is time-consuming, as it involves ensembling multiple transformers to enhance model performance. In contrast, our approach achieves comparable results with a significantly less amount of labeled data and using only one transformer. Our approach maintains high performance while being more resource-efficient by making the best use of the available data.

\begin{table}[hbt!]
\centering
\caption{Performance comparison between the proposed models and existing approaches.}
\label{tab:comparison}
\renewcommand{\arraystretch}{1.2} % Adjust row height
\begin{tabular}{l c c c c}
\toprule
\textbf{Method} & \textbf{STL} & \textbf{MTL} & \textbf{Train Size} & \textbf{Macro F1-score} \\
\midrule
Elkaref et al. \cite{elkaref2022guct}   & \ding{51} &        & 8,557   & 79.10 \\
Shapiro et al. \cite{shapiro2022alexu} & \ding{51} &        & 19,906  & 84.10 \\
Paula et al. \cite{de2022upv}          & \ding{51} &        & 8,557   & 82.70 \\
Mostafa et al. \cite{mostafa2022gof}   & \ding{51} &        & 8,557   & 85.17 \\
Ben Nessir et al. \cite{bennessir2022icompass} &        & \ding{51} & 12,473  & 83.90 \\
\midrule
\textbf{STL Model (Ours)} & \ding{51} &        & 8,557   & 85.31 \\
\textbf{MTL Model (Ours)} &        & \ding{51} & 3,336   & \textbf{85.42} \\
\bottomrule
\end{tabular}
\end{table}

\subsection{Ablation Study}
\vspace{2mm}\noindent\textbf{Emojis.}
Table \ref{tab_emojis} shows the performance of the proposed framework across all MTL and uncertainty approaches with different emoji settings. As shown in the table, the inclusion of emojis consistently leads to performance improvements compared with the same settings without emojis. This improvement can be attributed to the contribution of emojis that enable the model to capture additional semantic and contextual information. As shown in the table, 
the most substantial improvements are obtained with weighted emojis where more weights are assigned to a predefined list of emojis commonly associated with offensive speech. This indicates that emojis' contribution needs careful adjustment rather than being treated equally with other features.

\begin{table}[]
\centering
\caption{The effect of emojis on the performance of different MLT and uncertainty approaches. }
\label{tab_emojis}
\resizebox{\linewidth}{!}{\begin{tabular}{c|cccc|cccc|cccc}
\toprule
\multicolumn{1}{c|}{\multirow{2}{*}{Emojis}} & \multicolumn{4}{c|}{Equal}                                                                                          & \multicolumn{4}{c|}{Static}                                                                                       & \multicolumn{4}{c}{Dynamic}                                                                                        \\ \cmidrule{2-13} 
\multicolumn{1}{c|}{}                        & \multicolumn{1}{c}{None} & \multicolumn{1}{c}{Equal} & \multicolumn{1}{c}{Weighted} & \multicolumn{1}{c|}{Dynamic} & \multicolumn{1}{c}{None} & \multicolumn{1}{c}{Equal} & \multicolumn{1}{c}{Weighted} & \multicolumn{1}{c|}{Dynamic} & \multicolumn{1}{c}{None} & \multicolumn{1}{c}{Equal} & \multicolumn{1}{c}{Weighted} & \multicolumn{1}{c}{Dynamic} \\ \midrule
No                                            & 83.93                     & 83.13                      & 83.10                        & 82.32                        & 84.17                     & 83.99                      & 83.72                       & 82.83                        & 84.68                     & 83.09                      & 83.51                       & 82.75                        \\
Yes                                           & 84.08                     & 84.28                      & 84.41                       & 84.34                        & 83.93                     & 85.19                      & 84.02                       & 83.91                        & 84.46                     & 84.47                      & 85.09                       & 84.32                        \\
Weighted                                      & 83.97                     & 84.15                      & 85.07                      & 83.82                        & 83.19                     & 85.19                      & 83.66                       & 84.57                        & 84.01                     & \textbf{85.42}                      & 84.6                        & 84.00    \\ \bottomrule                   
\end{tabular}}
\end{table}

\vspace{2mm}\noindent\textbf{Number of Selected samples.} 
Table \ref{tab_no_samples} presents an analysis of the impact of different numbers of selected samples on the performance of various MLT approaches combined with different uncertainty methods with weighted emojis. The number of selected samples varies across 10, 20, 30, and 40 samples. As shown in the table, the highest score was obtained with only 10 samples selected per batch with dynamic weighting MTL and equal uncertainty. Generally, static uncertainty benefits from more samples, while other uncertainty settings show inconsistent trends. 
These insights suggest that fine-tuning strategies should be carefully tailored based on the chosen MLT and uncertainty estimation approach, rather than assuming that increasing sample size will always lead to performance gains. 

% \begin{sidewaystable}

\begin{table}[]
\centering
\caption{The performance of the MTL with different numbers of selected samples. }
\label{tab_no_samples}
\resizebox{\linewidth}{!}{\begin{tabular}{c|ccc|ccc|ccc}
\toprule  
\multicolumn{1}{c|}{\multirow{2}{*}{\#   Selected Samples}} & \multicolumn{3}{c|}{Equal}                                                              & \multicolumn{3}{c|}{Static}                                                           & \multicolumn{3}{c}{Dynamic}                                                            \\ \cmidrule{2-10} 
\multicolumn{1}{c|}{}                                      & \multicolumn{1}{c}{Equal} & \multicolumn{1}{c}{Weighted} & \multicolumn{1}{c|}{Dynamic} & \multicolumn{1}{c}{Equal} & \multicolumn{1}{c}{Weighted} & \multicolumn{1}{c|}{Dynamic} & \multicolumn{1}{c}{Equal} & \multicolumn{1}{c}{Weighted} & \multicolumn{1}{c}{Dynamic} \\ \midrule
10                                                          & 83.97                      & 85.07                       & 83.82                        & 85.19                      & 83.66                       & 84.57                        & \textbf{85.42}                      & 84.60                        & 84.00                           \\
20                                                          & 84.28                      & 83.85                       & 84.55                        & 83.78                      & 84.04                       & 82.65                        & 84.28                      & 84.51                       & 84.03                        \\
30                                                          & 84.33                      & 83.34                       & 83.99                        & 84.66                      & 83.78                       & 84.5                         & 83.07                      & 81.95                       & 84.03                        \\
40                                                          & 84.47                      & 84.51                       & 83.14                        & 84.37                      & 85.05                       & 83.92                        & 83.84                      & 85.03                       & 84.17   \\ \bottomrule

\end{tabular}}
\end{table}

\vspace{2mm}\noindent\textbf{Task weights.}
We evaluate in Table \ref{tab:tasks_weights} the performance of the MTL method with different weights assigned to each task’s loss. The offensive task is assigned the highest weight in all experiments, as it represents the main task, while the others serve as supporting tasks. As shown in the table, increasing the weight of the offensive task from 0.50 to 0.70 leads to noticeable improvements in overall performance across most uncertainty approaches, particularly under the Equal and Dynamic methods. The highest F1-score, 85.19\%, is achieved when the offensive task is assigned a weight of 0.70 using the Equal uncertainty approach. However, overemphasizing the offensive task beyond this point results in a decline in performance, likely due to underrepresenting the auxiliary tasks. These findings highlight the importance of carefully tuning task weights to maintain a balanced focus and enhance learning effectiveness in MTL models.

\begin{table}[ht]
\centering
\caption{The performance of the static MTL method with different task weights.}
\label{tab:tasks_weights}
\begin{tabular}{ccc|cccc}
\toprule
\multicolumn{3}{c|}{Task Weights} & \multicolumn{4}{c}{Uncertainty Approach} \\
\cmidrule(lr){1-3} \cmidrule(lr){4-7}
Offensive & Violent & Vulgar & None & Equal & Static & Dynamic \\
\midrule
0.50 & 0.25 & 0.25 & 84.47 & 84.30 & 82.39 & 84.44 \\
0.60 & 0.20 & 0.20 & 84.01 & 84.51 & 84.25 & 83.33 \\
0.70 & 0.15 & 0.15 & 83.97 & \textbf{85.19} & 83.66 & 84.57 \\
0.80 & 0.10 & 0.10 & 84.08 & 84.49 & 83.76 & 83.77 \\
\bottomrule
\end{tabular}
\end{table}
\section{Conclusion}

This study presents a novel framework that integrates MTL with uncertainty-based active learning to enhance offensive speech detection in Arabic social media text. By jointly modeling offensive, violent, and vulgar speech detection tasks and leveraging shared representations through MTL, the framework demonstrated improved data efficiency and classification performance. The integration of entropy-based uncertainty sampling enabled the selection of the most informative samples, reducing the dependency on large-scale labeled datasets. The proposed dynamic task weighting and entropy-based uncertainty sampling strategies further optimize the model's efficiency and accuracy. Our approach achieved a new state-of-the-art macro F1-score of 85.42\% on the OSACT2022 dataset while using significantly fewer training samples compared to existing methods. The inclusion of weighted emojis also proves beneficial, which captures additional semantic cues critical for accurate classification. These findings underscore the potential of combining MTL and active learning for resource-efficient and scalable offensive language detection in Arabic.

Despite its promising results, several challenges remain. The model exhibited difficulty in understanding implicit sarcasm, cultural references, and contextually nuanced language. Additionally, the generalizability of the model needs to be validated by applying it to other Arabic offensive language datasets or multilingual benchmarks.

\section*{Acknowledgments}
The authors would like to thank King Fahd University of Petroleum and Minerals (KFUPM) for supporting this work. In addition, we would like to thank the SDAIA-KFUPM Joint Research Center for Artificial Intelligence for supporting this work.

%Bibliography
%\bibliographystyle{unsrt}  
%\bibliography{references}  

\end{document}